\PassOptionsToPackage{english}{babel}

\documentclass[10pt,journal]{IEEEtran}
\usepackage[ngerman]{babel}
\usepackage[english]{babel}
\usepackage{lettrine}
\usepackage{graphicx}
\usepackage{float}
\usepackage{color}
\usepackage{cite}
\usepackage{amsmath}
\usepackage{array}
\usepackage{makecell}
\usepackage{ragged2e}
\usepackage{layouts}

\addto\captionsenglish{}
\newcolumntype{P}[1]{>{\centering\arraybackslash}p{#1}}
\newcolumntype{M}[1]{>{\centering\arraybackslash}m{#1}}

\def\BibTeX{{\rm B\kern-.05em{\sc i\kern-.025em b}\kern-.08em
    T\kern-.1667em\lower.7ex\hbox{E}\kern-.125emX}}

\begin{document}

\selectlanguage{english}

\title{Low Resolution Face Recognition Using a Two-Branch Deep Convolutional Neural Network Architecture}

\author{Erfan~Zangeneh,
        Mohammad~Rahmati,
        and~Yalda~Mohsenzadeh}

\markboth{}
{Erfan Zangeneh, Mohammad Rahmati, and Yalda Mohsenzadeh}

\selectlanguage{english}
\IEEEtitleabstractindextext{%
\begin{abstract}
\justifying{We propose a novel coupled mappings method for low resolution face recognition using deep convolutional neural networks (DCNNs). The proposed architecture consists of two branches of DCNNs to map the high and low resolution face images into a common space with nonlinear transformations. The branch corresponding to transformation of high resolution images consists of 14 layers and the other branch which maps the low resolution face images to the common space includes a 5-layer super-resolution network connected to a 14-layer network. The distance between the features of corresponding high and low resolution images are backpropagated to train the networks. Our proposed method is evaluated on FERET data set and compared with state-of-the-art competing methods. Our extensive experimental evaluations show that the proposed method significantly improves the recognition performance especially for very low resolution probe face images (11.4\% improvement in recognition accuracy). Furthermore, it can reconstruct a high resolution image from its corresponding low resolution probe image which is comparable with the state-of-the-art super-resolution methods in terms of visual quality.}
\end{abstract}

}

\maketitle
\IEEEdisplaynontitleabstractindextext

\section{INTRODUCTION}
\lettrine{\textbf{I}}{n} the past few decades, face recognition has shown promising performance in numerous applications and under challenging conditions such as occlusion \cite{jia2009support}, variation in in pose, illumination, and expression \cite{martinez2002recognizing}. While many face recognition systems have been developed for recognizing high quality face images in controlled conditions \cite{zhao2003face}, there are a few studies focused on face recognition in real world applications such as surveillance systems with low resolution faces \cite{pnevmatikakis2007far}. One important challenge in these applications is that high resolution (HR) probe images may not be available due to the large distance of the camera from the subject. Thus the performance of traditional face recognition systems which are developed for high quality images degrades considerably for these images with low resolution (LR) face regions \cite{turk1991face ,belhumeur1997eigenfaces, he2005face}. These LR face images typically have a size smaller than $32 \times 24$ pixels with an eye-to-eye distance about 10 pixels \cite{wang2014low}.

\begin{figure}
\begin{center}
  \includegraphics[width=8.5cm]{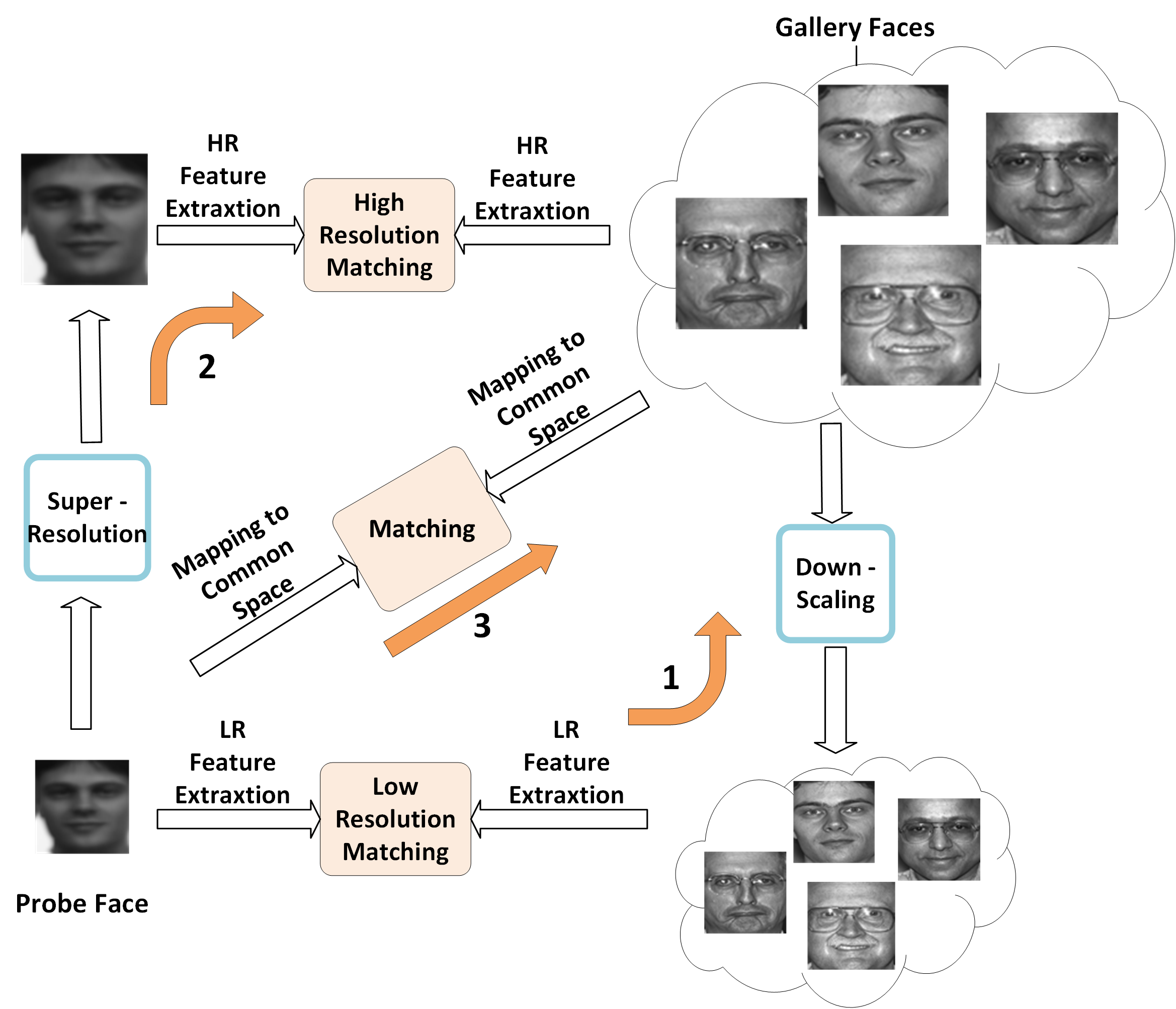}
\end{center}
  \caption{Three general approaches for low resolution face recognition.}\label{fig_allThreeApproaches}
\end{figure}

Here, we focus on addressing the problem of recognizing low resolution probe face images when a gallery of high quality images is available. There are three standard approaches to address this problem. 1) down sampling the gallery images to the resolution of the probe images and then performing the recognition. However, this approach is suboptimal because the additional discriminating information available in the high resolution gallery images is lost. 2) The second approach is to obtain higher resolution probe images from the low resolution images, which are then used for recognition. Most of these super-resolution techniques aim to reconstruct a good high resolution image in terms of visual quality and are not optimized for recognition performance \cite{simonyan2014very}. Some of the well known methods of this category are \cite{zou2012very, liu2007face, liu2005hallucinating, yang2010image} 3) Finally, the third approach simultaneously transforms both the LR probe and the HR gallery images into a common space where the corresponding LR and HR images are the closest in distance; \cite{hennings2008simultaneous, jian2015simultaneous, biswas2012multidimensional, zhou2011low} are the well known methods of this approach. Fig. \ref{fig_allThreeApproaches} summarizes the three general ways for low resolution face recognition (LR FR) problems.

In this paper, we use the third approach and propose a method that employs deep convolutional neural networks (DCNNs) to find a common space between low resolution and high resolution pairs of face images. Despite previous works that used linear equation as objective function to find two projection matrices, our work finds a nonlinear transformation from LR and HR to common space. In our proposed method, the distance of transformed low and high resolution images in the common space is used as an objective function to train our deep convolutional neural networks. Our proposed method also reconstructs good HR face images which are optimum for the recognition task. We evaluated the effectiveness of the proposed approach on the FERET database \cite{phillips2000feret}. Our results show the proposed approach improves the matching performance significantly as compared to other state-of-the-art methods in the low resolution face recognition and the improvement becomes more significant for very low resolution probe images.
The main contributions of this study can be summarized as:
\begin{itemize}
\item We proposed a novel nonlinear coupled mapping architecture using two deep convolutional neural networks to project the low and high resolution face images into a common space.
\item Our proposed method offers high recognition accuracy compared to other state-of-the-art competing methods especially when the probe image is extremely low resolution.
\item Our proposed coupled mappings method also offers high resolution version of the low resolution input image because of an embedded super-resolution CNN in its architecture.
\item Our proposed method needs much less space compared to the typical face recognition methods that use deep convolutional neural networks such as VGGnet\cite{parkhi2015deep}. This feature makes it applicable on regular systems with lower Memory.
\end{itemize}

In the following, we first review previous works presented in the domain of low resolution face recognition both super-resolution methods and coupled mapping methods in Section 2. In Section 3, we present our proposed method, network architecture and training procedure. Eventually, we present the experimental evaluation results in Section 4 and discussion and conclusion in Section 5.

\section{PREVIOUS WORKS}
In this section, we briefly review the related works in the literature of low resolution face recognition  and also introduce deep convolutional neural networks.

\textbf{Low resolution face recognition:}
 To resolve the mismatch between probe and gallery images, most of studies concentrated on super-resolution approaches. The aim of these approaches is to obtain a HR image from the LR input and then use the obtained HR image for recognition. Some super-resolution studies suggested using face priors for image reconstruction. The learning method proposed by Chakrabarti et al. \cite{chakrabarti2007super} for super-resolution uses kernel principal component analysis to derive prior knowledge about the face class. To achieve good reconstruction results, Liu et al. \cite{liu2007face} presented a two-step statistical modeling approach for hallucinating a HR face image from a LR input image. Baker \cite{baker2000hallucinating} also proposed a face hallucination method based on face priors. Freeman et al. \cite{freeman2000learning} trained a patch-wise Markov network as a super-resolution prediction model. Yang et al. \cite{yang2010image} used compressed sensing to reconstruct a super-resolution image from a low resolution input image. Zou et al. \cite{zou2012very} proposed a super-resolution method which clusters low resolution face images and then a projection matrix corresponding to the assigned cluster maps LR image into HR space. They proposed two separate projection matrices for optimal visualization and recognition purposes. In \cite{yang2010image}, the authors suggested a sparse coding method to find a representation of the LR input patch in terms of its neighboring image patches; then the same representation coefficients were used to reconstruct the target HR patch based on the corresponding neighboring HR patches.

The other category of works on LR FR is known as coupled mappings method. These methods learn the transformations using a training set consisting of HR images and LR images of the same subjects. Given training data, the goal is to find transformations which minimizes the distances between the transformed LR and HR feature vectors, $x_i^l$ and $x_i^h$, respectively. Most of coupled mappings methods use linear objective function as following \cite{li2010low}:
\begin{equation}\label{formule_formule1}
 J(W_L,W_H) = \sum_{i=1}^{n} \sum_{j=1}^{n}{||W_L^Tx_i^l - W_H^Tx_j^h||}^2P_{ij}
\end{equation}
\noindent where $n$ is the number of training images and ${\{x_i^h\}}_n^{i=1}$ and ${\{x_i^l\}}_n^{i=1}$ are corresponding extracted features of the HR and LR images, respectively. $W_L$ and $W_H$ denote the linear mappings of low resolution and high resolution feature vectors to the common space, respectively. $P$ is a $n \times n$ penalty weighting matrix that preserves the local relationship between data points in the original feature spaces and it is defined on the neighborhoods of the data points as follows:
\begin{equation}\label{formule_formule2}
P_{ij} = \bigg\{
\begin{tabular}{c c}
  $exp( -\frac{{||x_i^h - x_j^h||}^2}{\sigma ^ 2})$ & $j \in C(i)$ \\
  0 & otherwise
\end{tabular}
\end{equation}
\noindent Here, $C(i)$ contains the indices of $k$ nearest neighbors of $x_i^h$ in high resolution space and $\sigma$ is Gaussian function width which is defined as
  \begin{equation}\label{formule_formule3}
  \sigma = \frac{\alpha\sum_{i,j}{||x_i^h-x_j^h||^2}}{n^2}
  \end{equation}
\noindent where $\alpha$ is a scale parameter.Since it is assumed that HR feature space has more discriminative information, the goal of the above objective function is to find a common feature space similar to HR feature space. Finally after optimizing the above objective function, $W_L$ and $W_H$ will be found, and low and high resolution images can be transformed into common space with these mappings, respectively.
\begin{figure*}
\begin{center}
  \includegraphics[width=\textwidth]{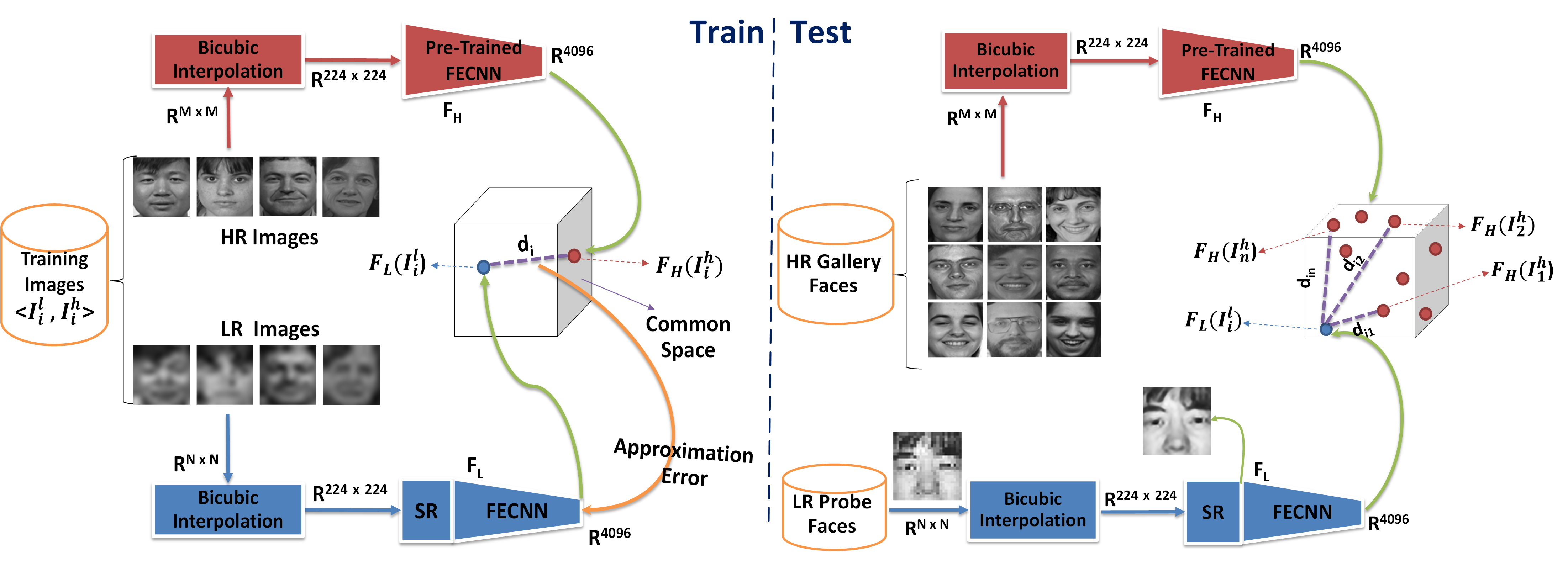}
\end{center}
  \caption{Overview of our method. M and N denote dimensions of HR and LR images, respectively, and $M>N$.}\label{fig_overviewOfOurMethod}
\end{figure*}
P. Hennings et al. \cite{hennings2008simultaneous} proposed a joint objective function which aims to optimize both super-resolution and face recognition. While this method improves the recognition accuracy compared to two-step methods, its optimization procedure is slow. Mainly because its optimization procedure has to be executed for each test image with respect to each enrollment.
In \cite{jian2015simultaneous}, the author used singular value decomposition (SVD) of face images in multi resolution to map low resolution images to high resolution space. Furthermore, the method improved both the hallucination and the recognition accuracy. Huang et al. \cite{huang2011super} proposed a method which finds a common space for low resolution probe and high resolution gallery images and an objective function that guarantees the discriminability in the new common space. Biswas et al. \cite{biswas2012multidimensional} used multidimensional scaling transformation learning to find both low resolution and high resolution projection matrices. The objective function of optimization problem enforces the same distance between low resolution and high resolution image pairs of a class in the common space as the distance of high resolution image pairs of that class. Huang et al. \cite{huang2011super} used canonical correlation analysis (CCA) to project low resolution and high resolution images into a common space where a low resolution image and its correspond high resolution image are as close as possible. Later, Ren et al. \cite{ren2012coupled} employed coupled nonlinear kernels to map the LR and HR face image pairs into an infinite common subspace. Also a coupled linear mappings method was presented in Zhou et al. \cite{zhou2011low} using the classical discriminant analysis. Shi et al. \cite{shi2015local} proposed an optimization objective function including three terms, associated with the LR/HR consistency, intraclass compactness and interclass separability. Zhang et al. \cite{zhang2015coupled} introduced coupled marginal discriminant mappings (CMDM) method. The method uses gaussian similarity between pairs of class-mean samples from HR images to construct intraclass similarity matrix. The interclass similarity matrix is defined by the gaussian similarity between sample pairs from HR images in the same class. Also, Zhang et al. \cite{zhang2015coupled} solved the coupled mappings problem as an eigen decomposition problem that helps to achieve good recognition performance when faces are occluded.
Mudunuri et al. \cite{mudunuri2016low} proposed a coupled mappings method that at first aligned faces by detecting eyes and then computed the SIFT descriptor of probe faces to transform them to a common space. Stereo matching cost function is then used to preserve distance in the transformed space across different illumination, pose and resolution.
\begin{figure*}
  \centering
  \includegraphics[width=\textwidth]{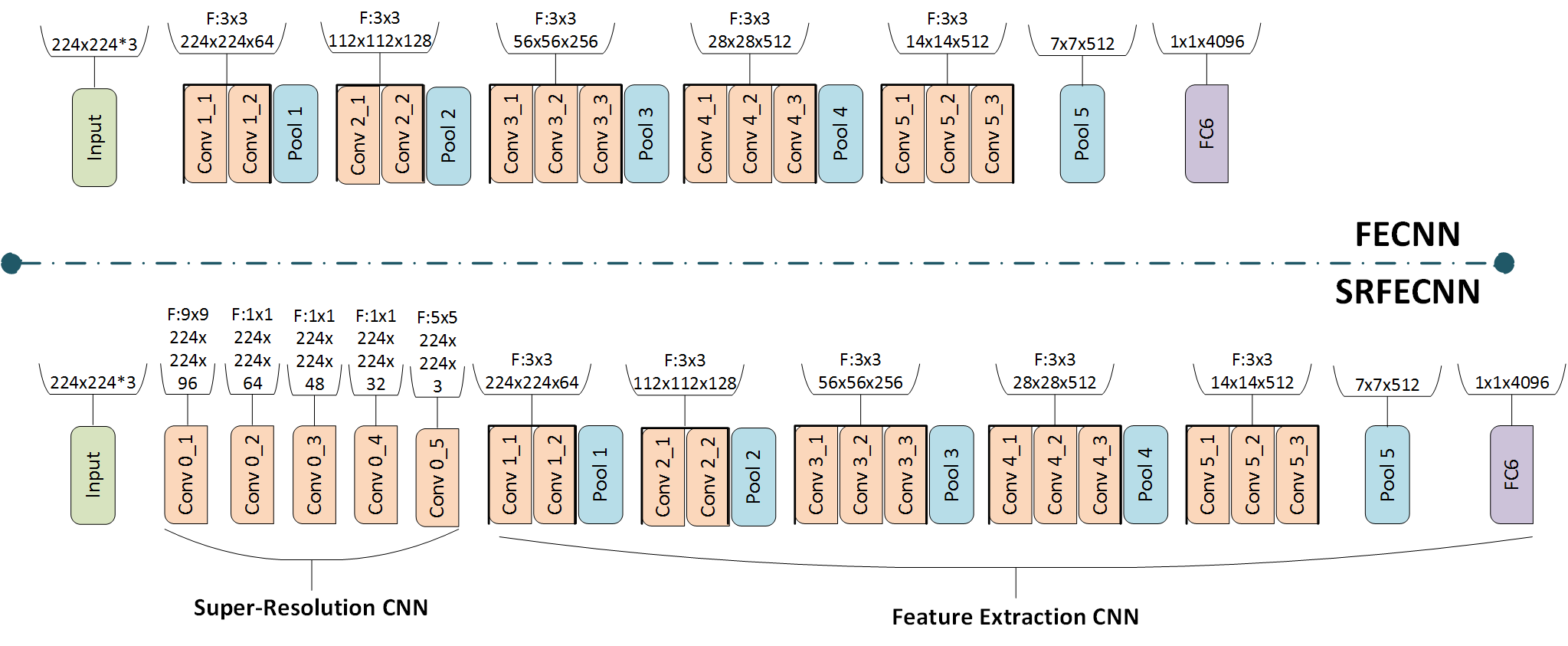}
  \caption{Architecture of two deep convolutional neural networks in two branches of our proposed method.}\label{fig_architecturesOFNets}
\end{figure*}

In summary, coupled mappings methods achieve better recognition performance than super-resolution methods, but these methods do not aim at reconstructing a high resolution image from the low resolution input image. On the other hand, the main objective of super-resolution methods is to reconstruct a high quality image for visualization purposes which may not necessarily offer better recognition accuracy.

\textbf{Deep convolutional neural networks:}
Although convolutional neural networks (CNN) were first presented three decades ago \cite{lecun1989backpropagation}, since introduction of AlexNet\cite{krizhevsky2012imagenet} in 2012, deep CNNs have become explosively popular especially due to their success in computer vision. The main elements that help to this popularity are\cite{dong2014learning}:
\begin{itemize}
  \item Participation of the best labs of top universities in computer vision challenges such as ILSVRC\cite{russakovsky2015imagenet} and PASCAL VOC\cite{Everingham15}
  \item Easy access to data with larger size such as ImageNet\cite{deng2009imagenet}.
  \item Introduction of more efficient activation functions like the rectified linear unit (ReLU)\cite{nair2010rectified}, and exponential linear unit (ELU) \cite{DBLP:journals/corr/ClevertUH15} which help DCNNs in faster convergence.
  \item Existence of modern GPU like NVIDIA TITAN black X, and also efficient deep learning frameworks such as Caffe\cite{jia2014caffe} and Tensorflow\cite{DBLP:journals/corr/AbadiBCCDDDGIIK16}.
\end{itemize}

In the next section, we propose a coupled mappings method using deep convolutional neural networks for nonlinear mapping to a common space. Our method similar to other successful methods that use deep convolutional neural networks, benefits from the above mentioned advantages. In addition to offering high recognition performance, the proposed method also produces high resolution images from low resolution input images.

\section{PROPOSED METHOD}
Due to the difficulty of solving a nonlinear optimization problem, objective functions in previous coupled mappings methods (as discussed in Section 2) were modeled with a linear transformation. However, a nonlinear transformation of low resolution and high resolution to a common space can possibly result in a better performance. Here, we propose a nonlinear coupled mappings approach which uses two deep convolutional neural networks (DCNNs) to extract features from low resolution probe images and high resolution gallery images and project them into a common space. We use gradient based optimization to minimize the distance between the mapped HR and LR image pairs in the common space with updating the weights of DCNN by backpropagation of the error.

Fig. \ref{fig_overviewOfOurMethod} shows the overview of our proposed architecture. In training phase, we use a training image set that contain pairs of low resolution and high resolution images of the same person which can vary in different images under different conditions of illumination, pose and expression (not necessarily the same image only with different resolutions). In the next section, we present the architecture of the proposed method in detail.
\subsection{Networks Architecture}
Our method has a two branch architecture that one of them projects high-resolution images to the common space and the other one maps low-resolution images into this common space. In our method we use a DCNN known as VGGnet \cite{simonyan2014very}. The most well-known configuration of this network has sixteen layers with thirteen convolutional layers and three fully connected layers. The last fully connected layer of VGGnet used for a specific classification task. In the top branch of our method (Fig. \ref{fig_overviewOfOurMethod}), we dropped out two last fully connected layers of this VGGnet and called it feature extraction convolutional neural network (FECNN). The input image of the top branch of our method is the high resolution image ($I_i^h$) that has to be in $224 \times 224$ dimensions (whenever input image size is different from $224 \times 224$, we use traditional bicubic interpolation method to obtain the required size). The output from the last layer is a feature vector with 4096 elements.

In the bottom branch of our method, we use a DCNN previously used for super-resolving low resolution images following by a second network which has a similar architecture as the network in the top branch. The first subnet has a similar architecture as DCNN that proposed by Dong et al. \cite{dong2014learning}, but we extended this architecture from three layers to five layers, although the authors show there is no difference between a three layer architecture and a five layer one in terms of visual quality of reconstructed images, we found increasing of layers from three to five improves the recognition performance of our method. We call the first subnet of our bottom branch super-resolution net (SRnet). The output of the first subnet is fed into the second subnet (FECNN). Therefore, the top branch net of our method consists of fourteen layers and the bottom branch includes nineteen layers as shown in Fig. \ref{fig_architecturesOFNets}. The input of bottom branch net is the low resolution image($I_i^l$) that has to be interpolated with the traditional interpolation method to the size of $224 \times 224$. Also, the output of SR subnet is an image with the size of $224 \times 224$.

As mentioned above, the FECNN net has the same architecture as VGGnet excluding the last two fully connected layers. Although the super resolution and feature extraction convolutional neural network (SRFECNN) has eighteen convolutional layers and one fully connected, the entire number of  weights used in SRFECNN is much less than VGGnet. Table \ref{table_weightsOfNet} shows all used weights for SRFECNN.
\begin{table}[h]
\caption {Number of used weights in layers of SRFECNN.} \label{table_weightsOfNet}
\begin{center}
\begin{tabular}{M{2.2cm} M{1.8cm} M{3.7cm}}
\Xhline{2\arrayrulewidth}
Layer set & Parameters & \ Number of weights\\
\hline
Conv0\_1& $F=9 \times 9 \newline Depth= 96$ & $3\times9\times9\times96 = 23328$\\
Conv0\_2& $F=1 \times 1 \newline Depth= 64$ & $96\times1\times1\times64 = 6144$\\
Conv0\_3& $F=1 \times 1 \newline Depth= 48$  & $64\times1\times1\times48 = 2928$\\
Conv0\_4& $F=1 \times 1 \newline Depth= 32$ & $48\times1\times1\times32 = 1536$\\
Conv0\_5& $F=5 \times 5 \newline Depth= 3$ & $32\times5\times5\times3 = 2400$\\
Conv1 (2 Convs)& $F=3 \times 3 \newline Depth= 64$ & $2(3\times3\times3\times64) = 3456$\\
Conv2 (2 Convs)& $F=3 \times 3 \newline Depth= 128$ & $2(64\times3\times3\times128) = 147456$\\
Conv3 (3 Convs)& $F=3 \times 3 \newline Depth= 256$ & $3(128\times3\times3\times256) = 884736$\\
Conv4 (3 Convs)& $F=3 \times 3 \newline Depth= 512$ & $3(256\times3\times3\times512) = 3538944$\\
Conv5 (3 Convs)& $F=3 \times 3 \newline Depth= 512$ & $3(512\times3\times3\times512) = 7077888$\\
FC6& $Depth = 4096$ & $7\times7\times512\times4096 = 102760448$\\
\hline
All Layers&  & $114449264 \approx 114M$\\
\Xhline{2\arrayrulewidth}
\end{tabular}
\end{center}
\end{table}
Even though our proposed SRFECNN includes eighteen convolutional layers, because of less number of fully connected layers compared to VGGnet, it has less number of weights than VGGnet ($141M$ weights). Thus in testing phase when we need to load SRFECNN weights on memory, our proposed method needs much less space than VGGnet. This is an important feature which makes our proposed method applicable on systems with lower memory.
\subsection{Common Subspace Learning}
We trained our network in three stages as summarized below:
\begin{itemize}
\item First, we used trained VGGnet on face dataset\cite{parkhi2015deep} and then dropped the last two fully connected layers, because they are specific to the classification task the network is trained on. We called this network pre-trained FECNN and used it in both top and bottom branches of our architecture.
\item In the second step, we trained the SRnet of the bottom branch with a database of high and low resolution face image pairs. The details of used datasets are presented in the experimental evaluation section.
\item The third step is the main training phase. We merged the two subnets namely SRnet and FECNN and a training database that contains pairs of low resolution and high resolution of same persons was fed into the bottom and top branches, respectively.
\end{itemize}
We considered the top branch FECNN net and the bottom branch SRFECNN as two nonlinear functions that project a high resolution image and low resolution image to a 4096 dimensional common space:
\begin{equation}\label{formule_formule4}
    \phi_i^h = F_H(I_i^h)
\end{equation}
\begin{equation}\label{formule_formule5}
    \phi_i^l = F_L(I_i^l)
\end{equation}
where $I_i^h \in R^{M \times M}$ and $I_i^l \in R^{N \times N}$ that $N < M$.
During this phase of training $F_H(I_i^h)$ was considered fixed and did not change, but  $F_L(I_i^l)$
was trained to minimize the distance between low and high resolution images of same subjects in the common space. With this aim, the distance was backpropagated into the bottom branch net (both FECNN and SRnet) as an error.

The main training procedure was repeated many times for all pairs of training images. We reduced learning rate of all layers to fine-tune the weights obtained in the first two training phases. However, the learning rate of first layers of FECNN is less than last layers of it, because in a specific problem, last layers of a DCNN have more discriminant information about the problem and the first layers of it have more general features that can change sparsely\cite{zeiler2014visualizing}.
\subsection{Reconstruct Input Image}
Additionally, our method can reconstruct a high resolution image from the low resolution probe image. First subnet of the bottom branch used for super-resolution to produce a high resolution face image from the low resolution probe face to feed into FECNN. In the test phase, after feeding low resolution probe image into the bottom net we can extract corresponding high resolution face image from the last layer of SRnet.
\subsection{Test Phase}
At first in the testing phase, all high resolution gallery images are fed to the top branch net and mapped into the common space and the probe image is fed into the bottom branch net. The label of probe image is determined by following formulae
  \begin{equation}\label{formule_formule6}
    Label(I_i^l) = Label(I_k^h)
  \end{equation}
\noindent where k determined by
  \begin{equation}\label{formule_formule7}
    k = \arg \min_j \{d_{i,j} \}_{j=1}^{N_G}
  \end{equation}
\noindent where $I_i^l$ is the low resolution probe image, $I_k^h$ is the $k^th$ high resolution gallery image and $N_G$ denotes number of high resolution face gallery images.
\section{EXPERIMENTAL EVALUATION}
The two primary tasks of face recognition are face identification and verification. In face identification, a query face is compared to the gallery face database to determine its identity. In face verification, the claimed identity of a query face is verified. In this section, all of the experiment results belong to face identification task. The experiments are designed to answer the following questions:
\begin{itemize}
\item How well does the proposed approach perform compared to the state-of-the-art
super-resolution methods and coupled mappings approaches?
\item How robust does the proposed approach perform across different resolutions of probe images?
\item How robust is the proposed approach to variations in expression, illumination, and age?
\item How does the proposed approach perform when the super-resolution subnet is excluded from the architecture?
\item How well does the proposed method reconstruct a high resolution face image from the low resolution probe one?
\item How is the convergence of the proposed approach in training phase effected when SR subnet is excluded from its architecture?
\end{itemize}
In order to demonstrate the effectiveness of our proposed method, we compare the face recognition performance of our method with state-of-the-art competing methods including one super-resolution (discriminative super-resolution (DSR) method \cite{zou2012very}) and three coupled mappings approaches (coupled locality preserving mappings (CLPM) \cite{li2010low}, nonlinear mappings on coherent features (NMCF) \cite{huang2011super}, and multidimensional scaling (MDS) \cite{biswas2012multidimensional} methods).
\subsection{Data Description}
\label{data_description}
In this section, we describe the datasets we used to train and evaluate our proposed method.

\textbf{Training dataset:} The details of datasets we used for training are presented in Table \ref{table_datasetsIntroduction} In total we used 45315 face images with variations in pose, expression, illumination and age for training. From FERET dataset \cite{FERET}, we used 10585 images in training and the rest (3541 images) in the evaluation phase.

\textbf{Evaluation dataset:} We carried out our evaluations on part of FERET \cite{phillips2000feret} face database. The FERET face data set contains 14126 face images from 1199 individuals. A subset of this data set including 3541 images is assigned for evaluation. This dataset includes four probe categories, each one assigned with a gallery set. All gallery face images are frontal. The four probe categories characteristics are explained below:
\begin{itemize}
\item The first probe category is called $FB$ and it includes 1195 frontal face images. Its gallery set contains 1196 frontal face images with different expressions.
\item The second probe category which is called $duplicate I$ contains all duplicate frontal images in the FERET database (722 images). The gallery is the same gallery as $FB$ containing 1196 images.
\item The third category is called $fc$ which includes 194 images taken on the same day, but with a different camera and illumination condition. The gallery is the same gallery as $FB$ containing 1196 images.
\item The fourth category called $duplicate II$ consists of duplicate probe images which are taken at least with one year difference with acquisition of corresponding gallery image (different age condition). The gallery for $duplicate II$ probes is a subset of the gallery for other categories containing 864 images.
\end{itemize}
\begin{table}[h]
\caption {List of datasets used for training and their description in terms of number of images and their variability in conditions such as E:expression, I:illumination, and P:pose. \textsuperscript{*} Please note that FERET dataset contains 14126 images and we used 10585 image for training, and the rest, 3541 images, for evaluation.} \label{table_datasetsIntroduction}
\begin{center}
\begin{tabular}{M{2.4cm} M{1.3cm} M{3.8cm}}
\Xhline{2\arrayrulewidth}
    Databases & Number of images & Highlights\\
    \hline
    300-W\cite{sagonas2013300} & 600 & in the wild, large variations in E\&I\&P  \\
    HELEN\cite{le2012interactive} & 2330 & in the wild, large variations in E\&I\&P ,and has occlusion \\
    IBUG\cite{sagonas2013300} & 135 & in the wild, large variations in E\&I\&P \\
    AFW\cite{zhu2012face} & 250 & in the wild, large variations in E\&I\&P   \\
    Georgia Tech Face Database\cite{georgiaTechDb} & 750 & large variations in E\&I\&P  \\
    LFW\cite{huang2007labeled} & 13233 & in the wild, large variations in E\&I\&P ,and scale \\
    UMIST\cite{graham1998characterising} & 564 & gray scale, and variations in pose and race  \\
    YALE B\cite{georghiades2001few} & 5760 & gray scale, and variation in P\&I \\
    AT\&T\cite{samaria1994parameterisation} & 400 & variation of time, eye glasses, and E\&I \\
    FERET\cite{FERET} & $14126^{*}$ & changes in appearance through time, and P\&I\&E \\
    CK+\cite{lucey2010extended} & 10708 & variation P\&E \\
\Xhline{2\arrayrulewidth}
\end{tabular}
\end{center}
\end{table}
\subsection{Training Phase }
We used the pre-trained VGGnet weights \cite{parkhi2015deep} and dropped the last two fully connected layers to construct our FECNN. Also, before training of our two branches architecture, we trained SRnet on the training datasets described in Table \ref{table_datasetsIntroduction}. For SRnet training, we first down-sampled faces from all of the training images to make the LR faces for the corresponding HR images in the dataset. The SRnet includes five convolutional layers and we trained the network with 45315 pairs of LR and HR face images. After training of FECNN and SRnet separately, we connected the pre-trained SR and FECNN subnets. Then we trained our proposed architecture using 45315 faces. In this main part of the training phase, we reduced learning rate of each layer in bottom branch to fine-tune the bottom net on training for coupled mappings purpose.
\subsection{Robustness Against Expression, Illumination and Age Variations}
In this experiment, we evaluated our proposed method on the four categories of FERET evaluation datasets described in Section \ref{data_description}. Since the $FB$ images have different expression conditions, the $fc$ set includes probe images with different illumination conditions and $duplicate II$ set contains probe images with different age conditions compared to the corresponding gallery images, we can evaluate the robustness of our proposed method against these variations as well. In this experiment, the HR face images with the size of $72 \times 72$ pixels are aligned with the positions of the two eyes. The LR images with size of $12 \times 12$ pixels are generated by the operation of down-sampling and smoothing on aligned HR face images.

\begin{figure*}
  \centering
  \includegraphics[width=\textwidth]{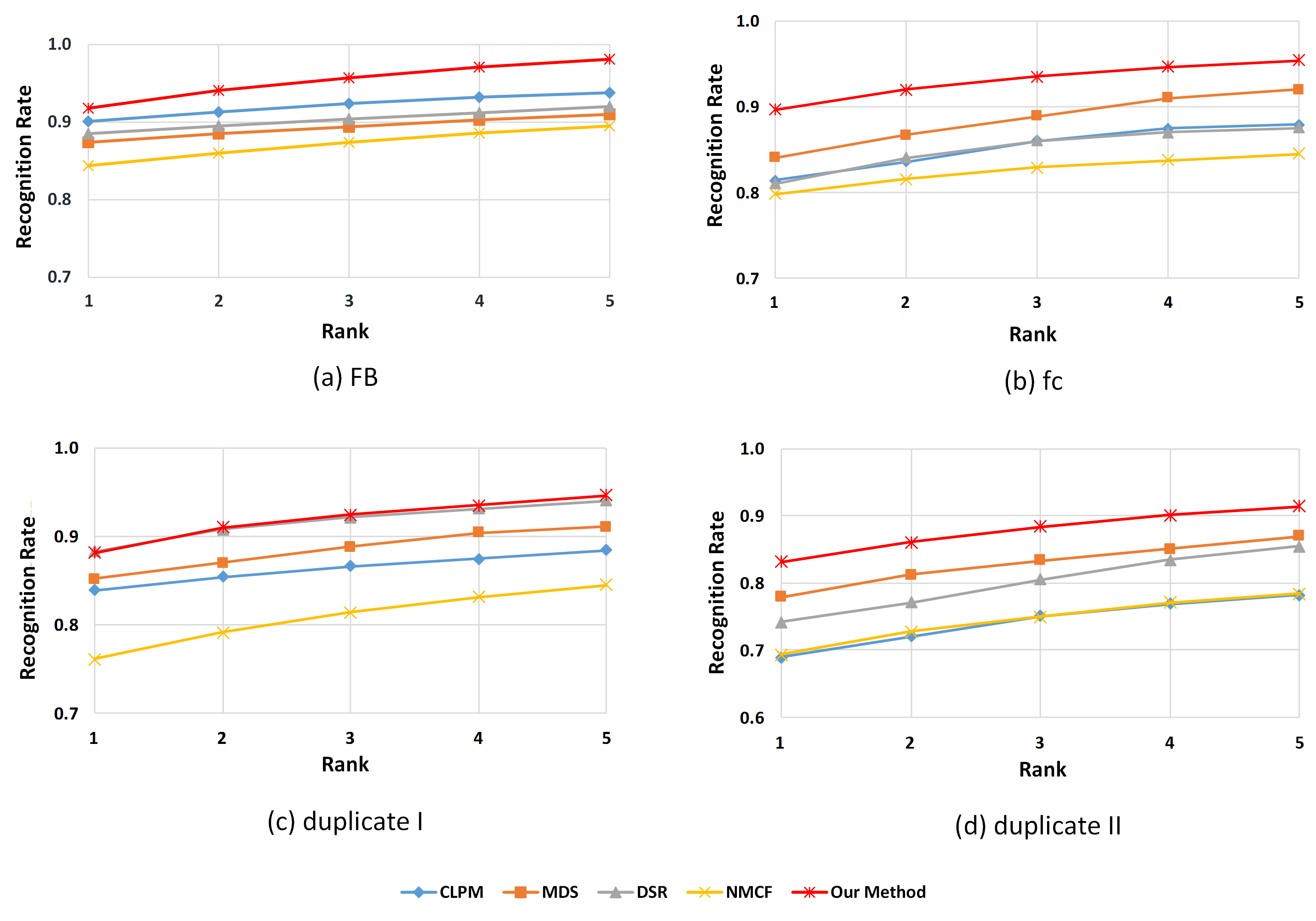}
  \caption{Comparison of our proposed method with CLPM\cite{li2010low}, MDS\cite{biswas2012multidimensional}, DSR\cite{zou2012very} and NMCF\cite{huang2011super} in terms of recognition rates. Cumulative match curves on (a) FB, (b) fc, (c) duplicate I, and (d) duplicate II datasets.}\label{fig_compareFB_fc_dupI_dupII}
\end{figure*}
Fig. \ref{fig_compareFB_fc_dupI_dupII} shows the cumulative match curve (CMC) for our method and four competing methods, DSR \cite{zou2012very}, MDS\cite{biswas2012multidimensional}, NMCF \cite{huang2011super}, and CLPM\cite{li2010low}. The cumulative match score for rank $k$ is a face identification measure which is defined as the recognition accuracy of the probe images when at least one of the $k$ nearest neighbors of the HR gallery images belongs to the same individual as the LR probe image. The results presented in Fig. \ref{fig_compareFB_fc_dupI_dupII} shows that the recognition performance of our method is significantly better than other state-of-the-art methods. Fig. \ref{fig_compareFB_fc_dupI_dupII}.a depicts the cumulative match curves on the $FB$ dataset. As we explained in Section\ref{data_description}, this dataset includes probe images different from gallery images only in terms of expression. The recognition accuracy of our proposed method in the rank 1 is 91.8\%, while the best performance of the competing methods belongs to CLPM\cite{li2010low} with 90.1\% recognition accuracy. Our proposed method outperforms the competing methods with 1.7\% difference. Fig. \ref{fig_compareFB_fc_dupI_dupII}.b depicts the CMC results on $fc$ dataset. The probe images in this dataset vary in illumination compared to gallery images. Our proposed method outperform competing methods across all ranks significantly. In rank 1, our method demonstrates an increase of 5.6\% compared to the best competing method on $fc$ dataset. This basically shows the efficiency of deep convolutional neural networks in feature extraction and generalization even in different illumination conditions. While the performance of our method is robust against the changes in illumination, the other competing methods performance drops significantly on $fc$ dataset compared to $FB$. $Duplicate I$ includes images in similar condition as the gallery, but with slightly expression variation. On this dataset, the performance of our method is close to the best competing method (DSR\cite{zou2012very}) (Fig. \ref{fig_compareFB_fc_dupI_dupII}.c). The $duplicate II$ contains probe images with different age condition compared to gallery images. Our proposed method outperforms the best competing method (here MDS\cite{biswas2012multidimensional}) on rank 1 with 5.2\% recognition accuracy (Fig. \ref{fig_compareFB_fc_dupI_dupII}.d). Again, this shows the robustness of our proposed method against variations in age.

\begin{table}[h]
\begin{center}
\caption {Comparison of rank 1 recognition accuracy across different probe image resolutions.} \label{table_probeResolution}
\begin{tabular}{ M{2cm}  M{1.2cm}  M{1.2cm}  M{1.2cm}  M{1.2cm} }
\Xhline{2\arrayrulewidth}
  &$6\times6$ &$12\times12$ &$24\times24$ &$36\times36$ \\
  \hline
  CLPMs\cite{li2010low}&64.4\% &90.1\% &93.4\% &95.2\% \\
  MDS\cite{biswas2012multidimensional}&57.3\% &87.4\% &90.2\% &92.2\% \\
  NMCF\cite{huang2011super}&60.3\% &84.4\% &88.4\% &91.1\% \\
  DSR\cite{zou2012very}&69.4\% &88.5\% &90\% &93\% \\
  Our Method&80.8\% &91.8\% &96.7\% &98.8\% \\
\Xhline{2\arrayrulewidth}
\end{tabular}
\end{center}
\end{table}
Taken together, our proposed method shows the best performance on $FB$, $fc$, and $duplicate II$ probe images and close to the best on $duplicate I$ dataset. Also our method shows robustness against variations in expression, illumination and age as shown in Fig. \ref{fig_compareFB_fc_dupI_dupII}.b and d.
\begin{figure}
  \centering
  \includegraphics[width=9cm]{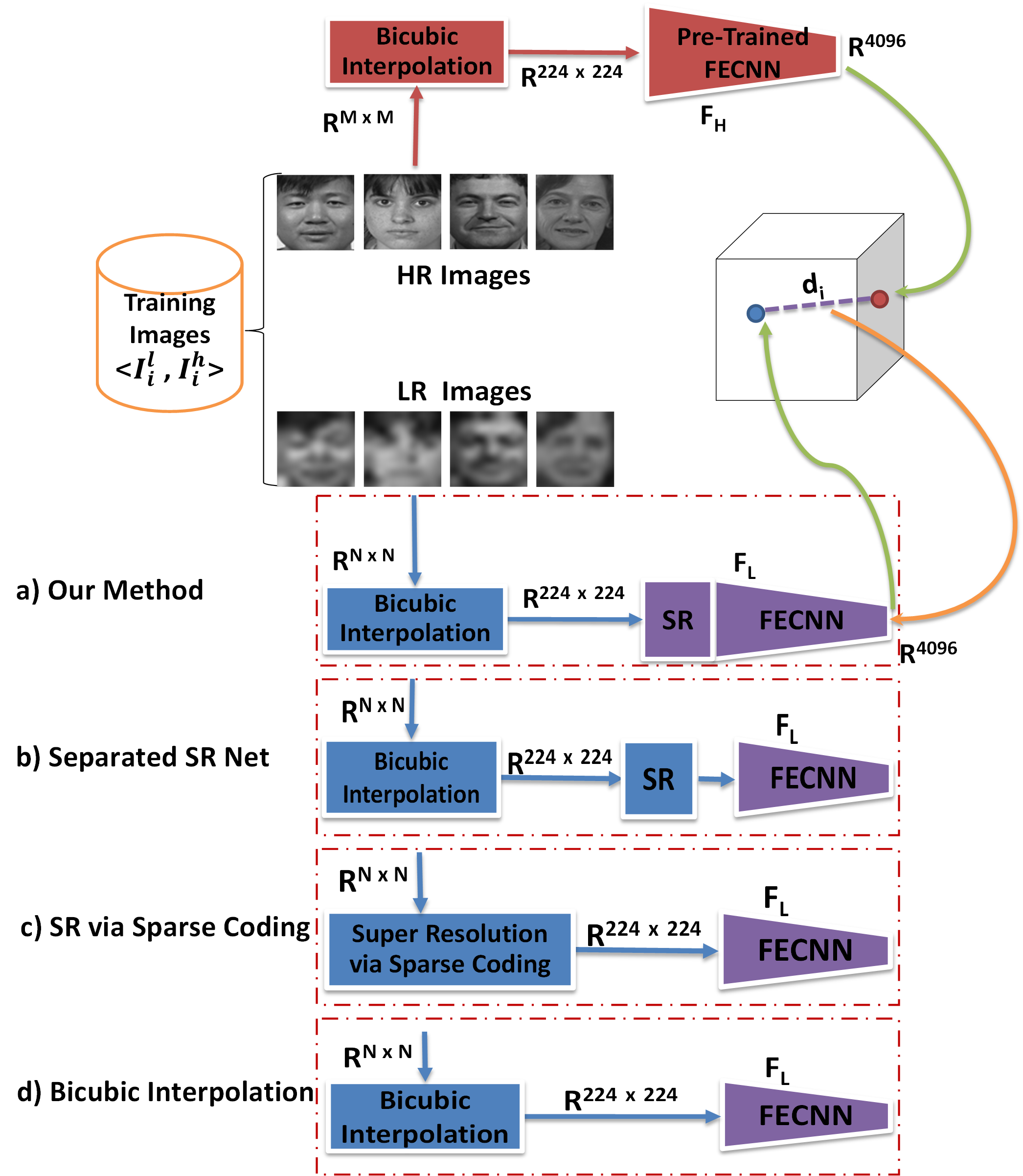}
  \caption{Configurations with different super-resolution modules. Modules with violet color are involved in training phase. a) Configuration of our method. b) SR subnet is separated from SRFECNN. c) Using sparse coding for SR \cite{yang2010image} d) Using only bicubic interpolation.}\label{fig_differentConfigurWithoutSR}
\end{figure}
\subsection{Evaluation on Different Probe Resolutions}
Here, we evaluated the effectiveness of our proposed method on probe images with very low resolutions. In this experiment, we compared the performance of our method with state-of-the-art methods on $FB$ probe dataset which all probe faces of this dataset are similar to gallery faces, but with slighly variation in expression. Thus appropriate to study the effect of variations in resolution. We considered four different resolutions, $6\times6$, $12\times12$, $24\times24$, and $36\times36$. Each time, we trained the SRnet separately on training data with reduced images resolutions and then connected the SRnet to FECNN and retrained the bottom branch of our proposed method on each resolution condition separately. Table \ref{table_probeResolution} shows the rank 1 recognition accuracy of our method compared to the competing methods on different resolution conditions evaluated on $FB$ set. As can be seen, our proposed method outperforms all the competing methods on all four resolution conditions. The most significant improvement (11.4\%) is on the very low resolution of $6\times6$ where our proposed method beats DSR\cite{zou2012very}, a method specifically proposed for the recognition of very low face images.
\begin{figure}
  \centering
  \includegraphics[width=8.5cm]{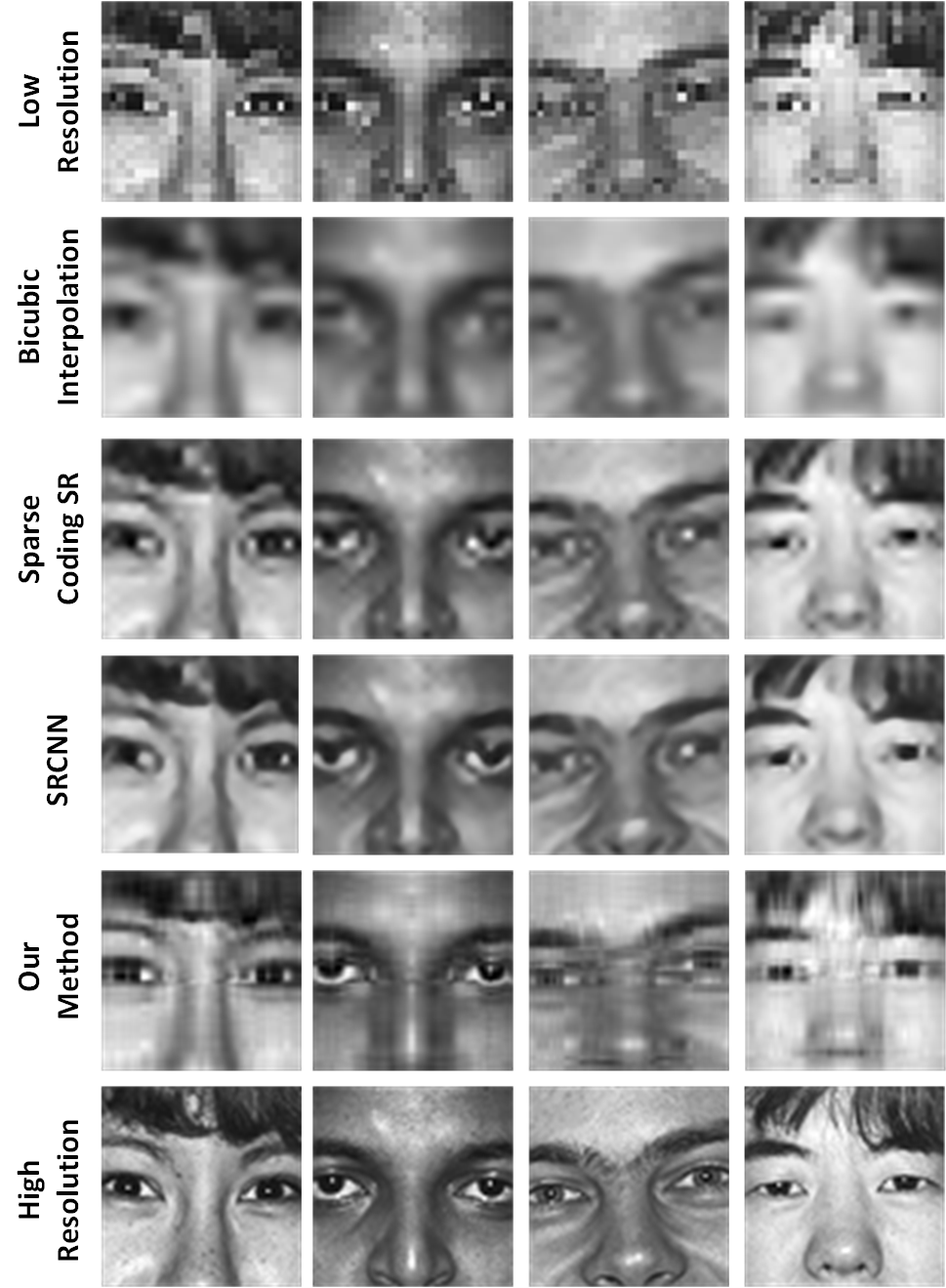}
  \caption{Reconstructed Faces by different configurations in Fig. \ref{fig_differentConfigurWithoutSR}.}\label{fig_reconstructedFaces}
\end{figure}
\subsection{The Role of SR Subnet}
\label{role_of_SR}
As explained, the bottom branch net is consist of two nets, SR and FECNN. In training phase, both SR and FECNN nets are involved in the main training phase. In this experiment, we aim to study the impact of using SRnet and also its fine-tuning on the recognition performance of our method.
\begin{figure*}
  \centering
  \includegraphics[width=\textwidth]{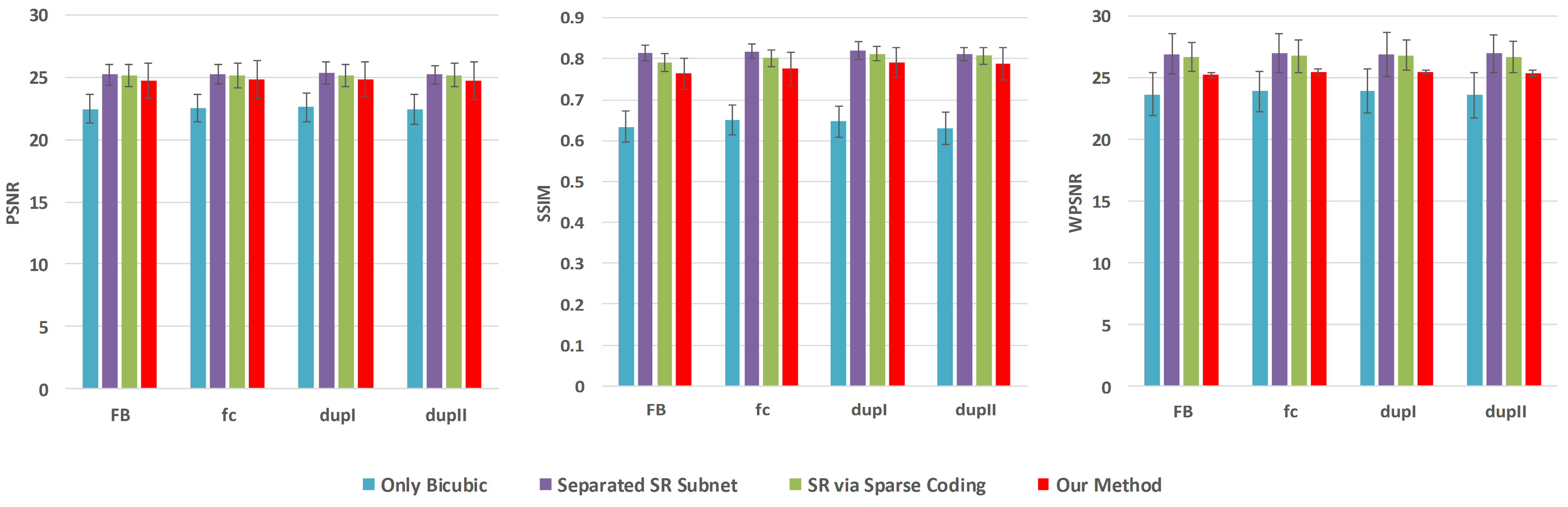}
  \caption{Visual quality comparison of reconstructed HR faces in terms of PSNR, SSIM and WPSNR, while scale factor is 3.}\label{fig_psnrSsimWpsnr}
\end{figure*}
\begin{table}[b]
\caption {Comparison of rank 1 recognition accuracy for different SR module configurations across different probe image resolutions.}
\label{table_accWithoutSR}
\centering
\begin{center}
\begin{tabular}{ M{3cm}  M{0.8cm}  M{0.8cm}  M{0.8cm} M{0.8cm} }
\Xhline{2\arrayrulewidth}
  &$6\times6$ &$12\times12$ &$24\times24$ &$36\times36$ \\
  \hline
  Only Bicubic&66.7\% &82.1\% &88.6\% &93.9\% \\
  Separated SR Subnet&75.4\% &89.7\% &95.3\% &97.9\% \\
  SR via Sparse Coding&73.9\% &88.4\% &94.1\%&97\% \\
  Our Method&80.8\% &91.8\% &96.7\% &98.8\% \\
\Xhline{2\arrayrulewidth}
\end{tabular}
\end{center}
\end{table}
Fig. \ref{fig_differentConfigurWithoutSR} shows three different configurations that we compared our proposed method with them. Our proposed method configuration is depicted in Fig. \ref{fig_differentConfigurWithoutSR}.a where both SR and FECNN subnet are trained during the main training phase. In the configuration shown in Fig. \ref{fig_differentConfigurWithoutSR}.b, SRnet is separated from FECNN in bottom branch, and in the main training phase weights of SRnet are kept fixed. The configuration shown in Fig. \ref{fig_differentConfigurWithoutSR}.c employs sparse coding \cite{yang2010image} method instead of the SRnet. Again only the FECNN is trained during the main training phase. The configuration illustrated in Fig. \ref{fig_differentConfigurWithoutSR}.d uses only a bicubic interpolation to map the low resolution input image to an image of size $224\times224$ and thus no super-resolution net is used. Therefore, in the training phase, only FECNN weights are updated. Table \ref{table_accWithoutSR} shows, the rank 1 recognition accuracy of the four different configurations (see Fig. \ref{fig_differentConfigurWithoutSR}). These results illustrate that using the SRnet in the configuration improves the performance (see the second row of Table \ref{table_accWithoutSR}). Furthermore, involving the SRnet in the main training phase improves the recognition performance considerably (our proposed method in Table \ref{table_accWithoutSR}). Especially, when the resolution of probe set is very low, the recognition performance of our method is considerably higher than other configurations.
Together, we can conclude the employment and training of SRnet improves the recognition performance of our proposed method architecture especially for probe images with very low resolutions.

\begin{figure}
  \centering
  \includegraphics[width=8.5cm]{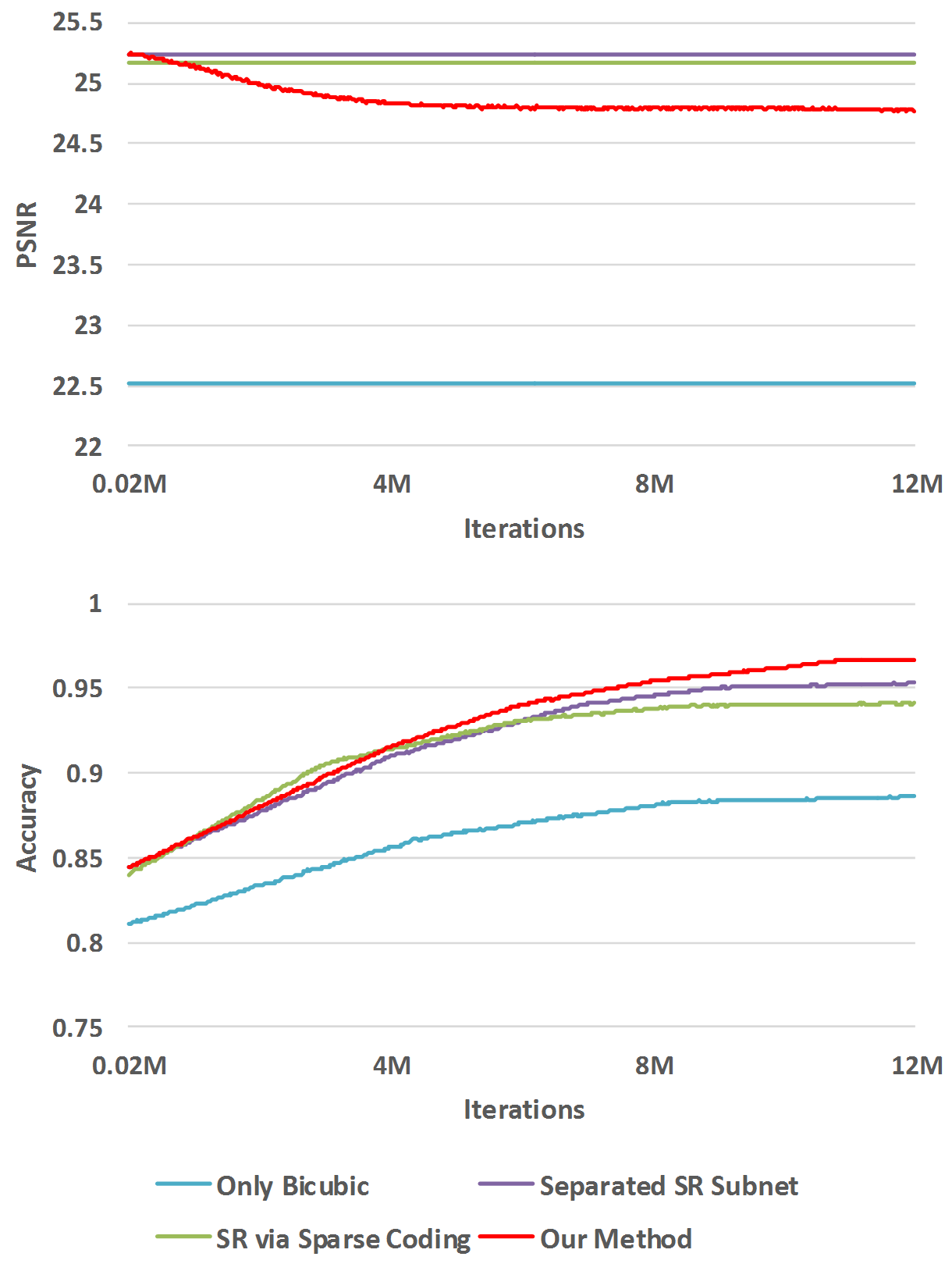}
  \caption{Changes in visual quality of reconstructed HR faces and recognition accuracy during training.}\label{fig_psnrVsAcc}
\end{figure}
\subsection{Evaluation on Reconstructed HR Face}
Despite other coupled mappings methods, our proposed method can also reconstruct a high resolution face from the low resolution one. In this experiment, we aim to evaluate our method in terms of high resolution face reconstruction. Here, we again compare the performance of our method with the three configurations introduced in Fig. \ref{fig_differentConfigurWithoutSR} in terms of visual quality of reconstructed face images. The size of low resolution images used in this section is $24 \times 24$ pixels.
Fig. \ref{fig_reconstructedFaces} shows some examples of reconstructed face images by each method. To compare visual enhancement of the four methods, peak signal to noise ratio (PSNR), structural similarity index (SSIM) and weighted peak signal to noise ratio (WPSNR\cite{voloshynovskiy1999stochastic}) metrics are used. As shown in Fig. \ref{fig_psnrSsimWpsnr}, when SRnet is separated from FECNN net, the reconstructed face images have the best visual quality and sparse coding is the second. Our method places in the third position in these results, however the differences between reconstructed face images by our method in comparison with the top two methods is small. As discussed in Section \ref{role_of_SR}, the recognition accuracy of our proposed method is much better compared to other configurations. This shows that the visual quality of super-resolved face images is compromised for better recognition performance in our proposed method. One interesting point is that the variance of PSNR and SSIM is higher for our method compared to other three competing methods. This shows that in some cases like the first two examples (on the left) in Fig. \ref{fig_reconstructedFaces}, the visual quality has improved while in others like the other two examples, the quality has degraded. In other words, the changes in SRnet has been in a direction to help the recognition performance eventually which is not necessarily in the direction of visual enhancement.
Fig. \ref{fig_psnrVsAcc} compares the changes of visual quality of reconstructed face images and recognition accuracy, during training for the four methods. As can be seen at the end of the training phase, our method achieved the best recognition performance but the third place in the visual quality of reconstructed face images.
\section{Conclusion}
In this paper, we presented a novel coupled mappings approach for the recognition of low resolution face images using deep convolutional neural networks. The main idea of our method is to use two DCNNs to transform low resolution probe and high resolution gallery face images into a common space where the distances between all faces belong to the same individual are closer than distances between faces belong to different persons. Our proposed method demonstrates significant improvement in recognition accuracy compared to the state-of-the-art coupled mapping methods (CLPM\cite{li2010low}, NMCF\cite{huang2011super}, MDS\cite{biswas2012multidimensional}) and super resolution method (DSR\cite{zou2012very}). Our proposed method shows significant improvement and robustness against variations in expression, illumination and age. Our method also outperforms competing methods across various resolutions of probe images and it shows even more considerable performance improvement (11.4\%) when applied on very low resolution images of $6 \times 6$ pixels. Our proposed method also offers HR image reconstruction which its visual quality is comparable with state-of-the-art super-resolution methods. The required space for our trained model is much less than the traditional deep convolutional neural networks trained for face recognition like VGGnet and thus our proposed low resolution face recognition method is applicable on systems with regular memory.

\nocite{*}
\bibliography{ref}
\end{document}